\documentclass{article}
\usepackage{ijcai22}

\usepackage{times}
\usepackage{soul}
\usepackage[hidelinks]{hyperref}
\usepackage[utf8]{inputenc}
\usepackage{graphicx}
\usepackage{amsmath}
\usepackage{amsthm}
\usepackage{algorithm}
\usepackage{algorithmic}
\usepackage{array}
\usepackage{subfig}
\usepackage{textcomp}
\usepackage{stfloats}
\usepackage{url}
\usepackage{booktabs}
\usepackage{amsfonts}
\usepackage{nicefrac} 
\usepackage{microtype} 
\usepackage{verbatim}
\usepackage{amssymb}
\usepackage{color}
\usepackage{multirow}
\usepackage{cases}
\usepackage{wrapfig}
\usepackage{lipsum}
\usepackage[numbers,sort&compress]{natbib}
\usepackage[usestackEOL]{stackengine}
\strutlongstacks{T}

\def\etal{\emph{et al.~}}
\def\ie{\emph{i.e.}}
\def\eg{\emph{e.g.}}

\title{Visual Similarity Attention}

\author{
Meng Zheng$^{1}$\and
Srikrishna Karanam$^1$\and
Terrence Chen$^{1}$\and
Richard J.~Radke$^{2}$\And
Ziyan Wu$^1$
\affiliations
$^1$United Imaging Intelligence, Cambridge MA, USA, 
$^2$Rensselaer Polytechnic Institute, Troy NY, USA\\
\emails
\{first.last\}@uii-ai.com,
rjradke@ecse.rpi.edu
}

\begin{document}

\maketitle

\begin{abstract}
While there has been substantial progress in learning suitable distance metrics, these techniques in general lack transparency and decision reasoning, \ie, explaining why the input set of images is similar or dissimilar. In this work, we solve this key problem by proposing the first method to generate generic visual similarity explanations with gradient-based attention. We demonstrate that our technique is agnostic to the specific similarity model type, \eg, we show applicability to Siamese, triplet, and quadruplet models. Furthermore, we make our proposed similarity attention a principled part of the learning process, resulting in a new paradigm for learning similarity functions. We demonstrate that our learning mechanism results in more generalizable, as well as explainable, similarity models. Finally, we demonstrate the generality of our framework by means of experiments on a variety of tasks, including image retrieval, person re-identification, and low-shot semantic segmentation. 
\end{abstract}

\section{Introduction}
We consider the problem of learning similarity predictors for metric learning and related applications. Given a query image of an object, our task is to retrieve, from a set of reference images, the object image that is most similar to the query image. This problem finds applications in a variety of tasks, including image retrieval \cite{kulis2013metric}, person re-identification (re-id) \cite{zheng2016person}, and even low-shot learning \cite{shaban2017one,chen2019hybrid}. There has been substantial recent progress in learning distance functions for these similarity learning applications \cite{BIER_ICCV17,AttentionBasedEF_ECCV18,msLoss_CVPR19,FastAP_CVPR19}.


Existing deep similarity predictors are trained in a distance learning fashion so that the features of same-class data points are close to each other in the learned embedding, while features of data from other classes are further away. Consequently, most techniques distill this problem into optimizing a ranking objective that respects the relative ordinality of pairs, triplets, or even quadruplets \cite{law2013quadruplet} of training examples. These methods are characterized by the specificity of how the similarity model is trained, e.g., data (pairs, triplets etc.) sampling \cite{wu2017sampling}, sample weighting \cite{zheng2019hardness}, and adaptive ranking \cite{rippel2015metric}, among others. However, a key limitation of these approaches is their lack of decision reasoning, \ie, explanations for why the model predicts the input set of images is similar or dissimilar. As we demonstrate in this work, our method not only offers model explainability, but such decision reasoning can also be infused into the model training process, in turn helping bootstrap and improve the generalizability of the trained similarity model.

\begin{figure}[t]
	\centering
	\includegraphics[draft=false,width=\linewidth]{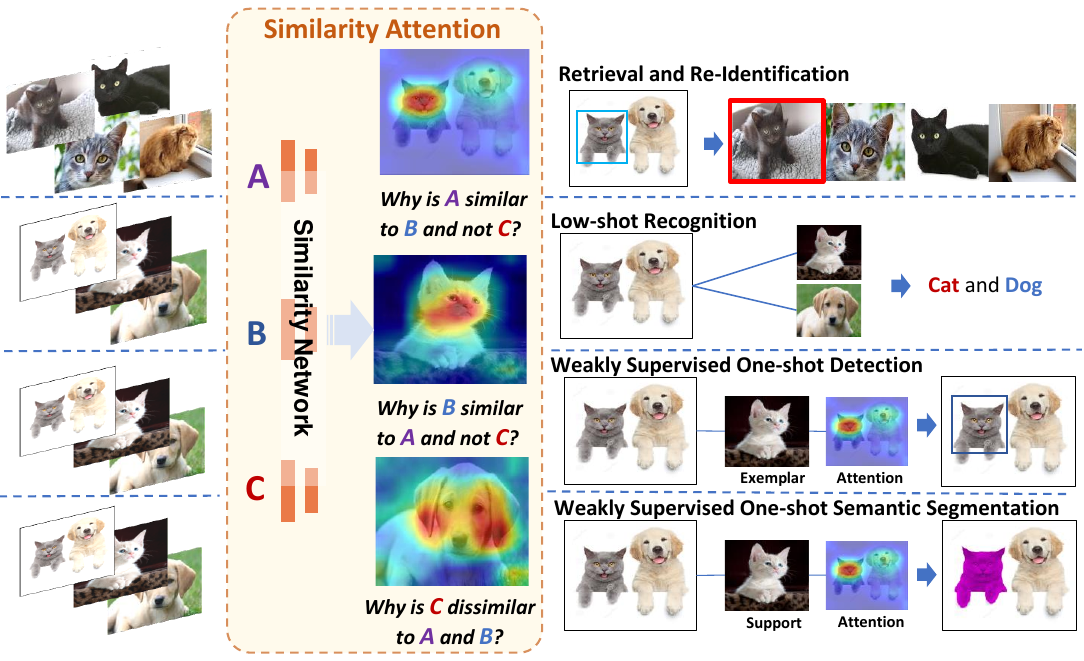}%
	\caption{Illustration of the proposed visual similarity explanation method and its applicability to different applications.} 
	\vspace{-1em}
	\label{fig:multipleApplications}
\end{figure}

Recent developments in CNN visualization \cite{mahendran2015understanding,GradCAM_ICCV17} have led to a surge of interest in visual explainability. Some methods \cite{GAIN_CVPR18,wang2019sharpen} enforce attention constraints using gradient-based attention \cite{GradCAM_ICCV17}, resulting in improved attention maps as well as downstream model performance. These techniques essentially ask: \textsl{where is the object in the image?} By design, this limits their applicability to scenarios involving object categorization. On the other hand, in this paper, we ask the question: \textsl{what makes image A similar to image B but dissimilar to image C?} (see Figure~\ref{fig:multipleApplications}). While existing works can explain classification models, their extensions to generating such visual similarity explanations is not trivial. A principled answer to this question will help explain models that predict visual similarity, which is what we address in our work.


To this end, we propose a new technique to generate CNN attention directly from similarity predictions. Note that this is substantially different from GradCAM-inspired \cite{GradCAM_ICCV17} existing work \cite{GAIN_CVPR18,wang2019sharpen,zheng2019re} where an extra classification module is needed to compute the network attention. Instead, our proposed method generates visual attention from feature vectors (produced by any CNN with fully-connected units) used to compute similarity (distance), thereby resulting in a flexible and generic scheme that can be used in conjunction with any feature embedding network. Furthermore, we show that the resulting similarity attention can be modeled as the output of a differentiable operation, thereby enabling its use in model training as an explicit trainable constraint, which we empirically show improves model generalizability. A key feature of our proposed technique is its generality, evidenced by two characteristics we demonstrate. First, our design is not limited to a particular type of similarity learning architecture; we show applicability to and results with three different types of architectures: Siamese, triplet, and quadruplet. Next, we demonstrate the versatility of our framework (Figure~\ref{fig:multipleApplications} shows a summary) in addressing problems different from image retrieval (\eg, low-shot semantic segmentation) by exploiting its decision reasoning functionality to discover (application-specific) regions of interest. 

To summarize, our key contributions include:
\begin{itemize}
    \item To the best of our knowledge, we present the first gradient-based technique, similarity attention, to generate visual explanations from generic similarity metrics, equipping similarity models with visual explainability.
    \item Our proposed method only requires feature vectors (produced by any CNN with fully-connected units) to generate visual attention, thereby extensible in principle to any feature embedding CNN model.  
    \item We show how the proposed similarity attention can be formulated into trainable constraints, resulting in a new similarity mining learning objective and enabling similarity-attention-driven learning mechanisms for training similarity models with improved generalizability.
    \item We demonstrate the versatility of our proposed framework by a diverse set of experiments on a variety of tasks (\eg, image retrieval, person re-id and low-shot semantic segmentation) and similarity model architectures. 
\end{itemize}

\vspace{-.5em}
\section{Related Work}
Our work is related to both the metric learning and visual explainability literature. In this section, we briefly review closely-related methods along these directions respectively. \\
\indent \textbf{Learning Distance Metrics.} Metric learning approaches attempt to learn a discriminative feature space to minimize intra-class variations, while also maximizing the inter-class variance. Traditionally, this translated to optimizing learning objectives based on the Mahalanobis distance function or its variants \cite{KISSME_CVPR12,LOMO_XQDA_CVPR15}. Much recent progress with CNNs has focused on developing novel objective functions or data sampling strategies. Wu \etal \cite{wu2017sampling} demonstrated the importance of careful data sampling, developing a weighted data sampling technique that resulted in reduced bias, more stable training, and improved model performance. Harwood \etal \cite{SmartMF_ICCV17} showed that a smart data sampling procedure that progressively adjusts the selection boundary in constructing more informative training triplets can improve the discriminability of the learned embedding. Substantial effort has also been expended in proposing new objective functions for learning the distance metric. Some recent examples include the multi-class N-pair \cite{Npairmc_NIPS16}, lifted structured embedding \cite{LiftStruct_CVPR15}, and proxy-NCA \cite{ProxyNCA_ICCV17} losses. The goal of these and related objective functions is essentially to explore ways to penalize training data samples (pairs, triplets, quadtruplets, or even distributions \cite{rippel2015metric}) so as to learn a discriminative embedding. In this work, we take a different approach. Instead of just optimizing a distance objective, we explicitly consider and model network attention during training. This leads to two key innovations over existing work. First, we equip our trained model with decision reasoning functionality. Second, by means of trainable attention, we guide the network to discover local image regions that contribute the most to the final decision, thereby improving model generalizability.\\
\indent \textbf{Learning Visual Explanations.} Dramatic performance improvements of vision algorithms driven by black-box CNNs have led to a recent surge in attempts \cite{mahendran2015understanding,CAM_CVPR16,residualAtt_CVPR17,GradCAM_ICCV17,AttentionIA_NIPS17,ABN_CVPR19} to interpret model decisions. To date, most CNN visual explanation techniques fall into either response-based or gradient-based categories. Class Activation Map (CAM) \cite{CAM_CVPR16} used an additional fully-connected unit on top of the original deep model to generate attention maps, thereby requiring architectural modification during inference and limiting its utility. Grad-CAM \cite{GradCAM_ICCV17}, a gradient-based approach, solved this problem by generating attention maps using class-specific gradients of predictions with respect to convolutional layers. There has been several works took a step forward, \eg \cite{GAIN_CVPR18, wang2019sharpen, zheng2019re} use the attention maps to enforce trainable attention constraints, demonstrating improved model performance. These aforementioned gradient-based techniques all require a well-trained classifier for generating visual explanations and reply on application-specific assumptions.
A few recent examples of attempts to visually explain similarity models include Plummer \etal \cite{whymatch_2019} and Chen \etal \cite{Chen_WACV20}. While Plummer \etal \cite{whymatch_2019} needs attribute labels coupled with an attribute classification module and a saliency generator to generate explanations, and Chen \etal \cite{Chen_WACV20} adopts a two-stage pipeline which first generates gradients by sampling training data tuples and then transfers gradient weight from training to testing by nearest neighbor search, our method is more \textsl{generic} that it does not need extra labels/additional learning modules as in \cite{whymatch_2019}, or training data access and weights transfer as in \cite{Chen_WACV20}. Our proposed algorithm can generate similarity attention from any similarity measure, and additionally, can enforce trainable constraints using the generated similarity attention. Our design leads to a flexible technique and generalizable model that we show competitive results in areas ranging from metric learning to low-shot semantic segmentation.


\vspace{-.5em}
\section{Proposed Method}
Given a set of $N$ labeled images $\{(\mathbf{x}_{i},y_{i})\}, i=1,\ldots,N$ each belonging to one of $k$ categories, where $\mathbf{x} \in \mathbb{R}^{H\times W\times c}$, and $y \in \{1,\ldots,k\}$, we aim to learn a distance metric to measure the similarity between two images $\mathbf{x}_{1}$ and $\mathbf{x}_{2}$. Our key innovation includes the design of a flexible technique to produce similarity model explanations by means of CNN attention, which we show can be used to enforce trainable constraints during model training. This leads to a model equipped with similarity explanation capability as well as improved model generalizability. In Section \ref{sec:SA}, we first briefly discuss the basics of existing similarity learning architectures followed by our proposed technique to learn similarity attention, and show how it can be easily integrated with existing networks. In Section~\ref{sec:SA_mining}, we discuss how the proposed mechanism facilitates principled attentive training of similarity models with our new similarity mining learning objective.

\subsection{Similarity attention}
\label{sec:SA}
Traditional similarity predictors such as Siamese or triplet models are trained to respect the relative ordinality of distances between data points. For instance, given a training set of triplets $\{(\mathbf{x}^{a}_{i},\mathbf{x}^{p}_{i},\mathbf{x}^{n}_{i})\}$, where $(\mathbf{x}^{a}_{i},\mathbf{x}^{p}_{i})$ have the same categorical label while $(\mathbf{x}^{a}_{i},\mathbf{x}^{n}_{i})$ belong to different classes, a triplet similarity predictor learns a $d-$dimensional feature embedding of the input $\mathbf{x}$, $f(\mathbf{x}) \in \mathbb{R}^{d}$, such that the distance between $f(\mathbf{x}^{a}_{i})$ and $f(\mathbf{x}^{n}_{i})$ is larger than that between $f(\mathbf{x}^{a}_{i})$ and $f(\mathbf{x}^{p}_{i})$ (within a predefined margin $\alpha$).

Starting from such a baseline predictor (we choose the triplet model for all discussion here, but later show variants with Siamese and quadruplet models as well), our key insight is that we can use the similarity scores from the predictor to generate visual explanations, in the form of visual attention maps \cite{GradCAM_ICCV17}, for why the current input triplet satisfies the triplet criterion w.r.t the learned feature embedding $f(\mathbf{x})$. As an example of our final result, see Figure~\ref{fig:multipleApplications}, where we note our model is able to highlight common (cat) face region in the anchor (A) and the positive (B) image, whereas we highlight the corresponding face and ears region for the dog image (negative, C), illustrating why this current triplet satisfies the triplet criterion. This is what we refer to by \textsl{similarity attention}: the ability of the similarity predictor to automatically discover local regions in the input that contribute the most to the final decision (in this case, satisfying the triplet condition) and visualize these regions by means of attention maps. 

\begin{figure}[!h]
	\centering
	\vspace{-1.5em}
	\includegraphics[draft=false,width=\linewidth]{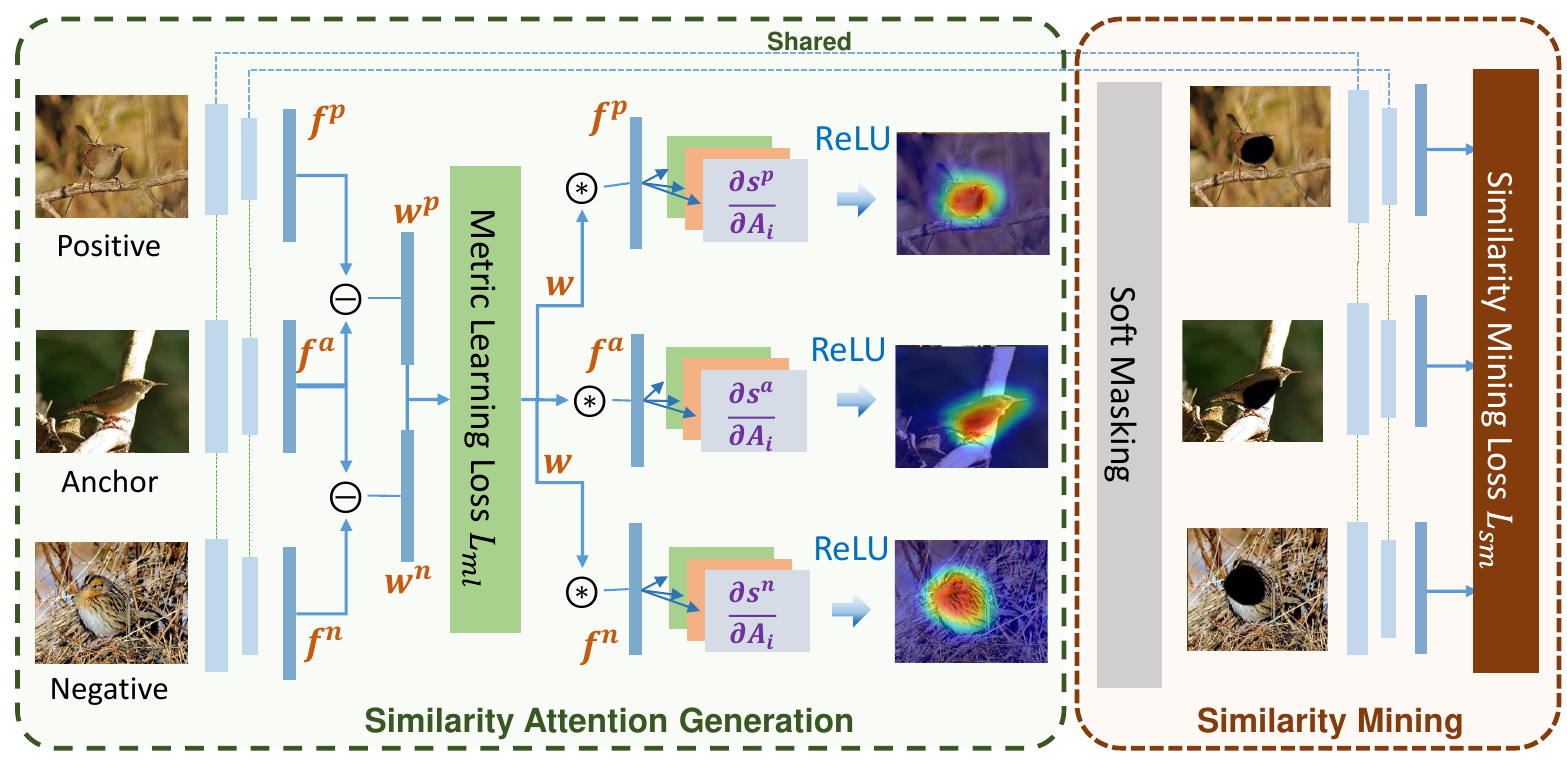}%
	\caption{Pipeline of proposed similarity attention and similarity mining techniques.} 
	\vspace{-1.2em}
	\label{fig:SA_mining}
\end{figure}

Note that our idea of generating network attention from the similarity score is different from existing work \cite{GradCAM_ICCV17,zheng2019re}, in which an extra classification module and the classification probabilities are used to obtain attention maps. In our case, we are not limited by this requirement of needing a classification module. Instead, as we discuss below, we compute a similarity score directly from the feature vectors (\eg, $f(\mathbf{x}^{a}_{i})$, $f(\mathbf{x}^{p}_{i})$, and $f(\mathbf{x}^{n}_{i})$), which is then used to compute gradients and obtain the attention map. A crucial advantage with our method is that this results in a flexible and generic scheme, that can be used to visually explain virtually any feature embedding network.\\
\indent An illustration of our proposed similarity attention generation technique is shown in Figure \ref{fig:SA_mining} (green dotted rectangle). Given a triplet sample $(\mathbf{x}^{a},\mathbf{x}^{p},\mathbf{x}^{n})$, we first extract feature vectors $\mathbf{f}^{a}$, $\mathbf{f}^{p}$, and $\mathbf{f}^{n}$ respectively. Note that all the feature vectors are normalized to have $l_{2}$ norm equal to 1. Ideally, a perfectly trained triplet similarity model must result in $\mathbf{f}^{a}$, $\mathbf{f}^{p}$, and $\mathbf{f}^{n}$ satisfying the triplet criterion. Under this scenario, local differences between the images in the image space will roughly correspond to proportional differences in the feature space as well. Consequently, there must exist some dimensions in the feature space that contribute the most to this particular triplet satisfying the triplet criterion, and we seek to identify these elements in order to compute the attention maps. To this end, we compute the absolute differences and construct the weight vectors $\mathbf{w}^{p}$ and $\mathbf{w}^{n}$ as $\mathbf{w}^{p}=\mathbf{1}-|\mathbf{f}^{a}-\mathbf{f}^{p}|$ and $\mathbf{w}^{n}=|\mathbf{f}^{a}-\mathbf{f}^{n}|$. 

With $\mathbf{w}^{p}$, we seek to highlight the feature dimensions that have a small absolute difference value (\eg, for those dimensions $t$, $\mathbf{w}^{p}_{t}$ will be closer to 1), whereas with $\mathbf{w}^{n}$ we seek to highlight the feature dimensions with large absolute differences. Given $\mathbf{w}^{p}$ and $\mathbf{w}^{n}$, we construct a single weight vector $\mathbf{w}=\mathbf{w}^{p} \odot \mathbf{w}^{n}$ ($\odot$ denotes element-wise product operation). With $\mathbf{w}$, we will obtain a higher weight with feature dimensions that have a high value in both $\mathbf{w}^{p}$ and $\mathbf{w}^{n}$. To further understand this intuition, let us consider a simple example. If the first feature dimension $f^{a}(1)=0.80$ and $f^{p}(1)=0.78$, then this first dimension is important for the anchor to be close to the positive. In this case, the first dimension of the corresponding weight vector $w^{p}(1)=(1-|0.80-0.78|)=0.98$, which is a high value, quantifying the importance of this particular feature dimension for the anchor and positive to be close. 
Given these high-value dimensions, we identify all such important dimensions common across both $\textbf{w}^{p}$ and $\textbf{w}^{n}$ with the single weight vector $\mathbf{w}$. In other words, we focus on elements that contribute the most to (a) the positive feature pair being close, and (b) the negative feature pair being further away. This way, we identify dimensions in the feature space that contribute the most to $\mathbf{f}^{a}$, $\mathbf{f}^{p}$, and $\mathbf{f}^{n}$ satisfying the triplet criterion. We now use these feature dimensions to compute network attention for the current image triplet $(\mathbf{x}^{a},\mathbf{x}^{p},\mathbf{x}^{n})$.

Given $\mathbf{w}$, we compute the dot product of $\mathbf{w}$ with $\mathbf{f}^{a}$, $\mathbf{f}^{p}$, and $\mathbf{f}^{n}$ to get the sample scores $s^{a}=\mathbf{w}^{T}\mathbf{f}^{a}$, $s^{p}=\mathbf{w}^{T}\mathbf{f}^{p}$, and $s^{n}=\mathbf{w}^{T}\mathbf{f}^{n}$ for each image $(\mathbf{x}^{a},\mathbf{x}^{p},\mathbf{x}^{n})$ respectively. We then compute the gradients of these sample scores with respect to the image's convolutional feature maps to get the attention map. Specifically, given a score $s^{i}, i\in \{a,p,n\}$, the attention map $\mathbf{M}^{i} \in \mathbb{R}^{m\times n}$ is determined as:
\vspace{-.5em}
\begin{equation}
\mathbf{M}^{i}=\text{ReLU}\left(\sum_k \alpha_{k}\mathbf{A}_{k}\right)
    \label{eq:mainSimilarityAttentionMap}
\end{equation}

where $\mathbf{A}_{k} \in \mathbb{R}^{m\times n}$ is the $k^{th} (k=1,\ldots,c)$ convolutional  feature channel (from one of the intermediate layers) of the convolutional feature map $\mathbf{A} \in \mathbb{R}^{m\times n \times c}$ and $\alpha_{k}=\text{GAP}\left(\frac{\partial s^{i}}{\partial \mathbf{A}_{k}}\right)$. The \text{GAP} operation is the same global average pooling operation described in GradCAM \cite{GradCAM_ICCV17}.

\subsubsection{Extensions to other architectures}

Our proposed technique to generate similarity attention is not limited to triplet CNNs and is extensible to other architectures as well. Here, we describe how to generate our proposed similarity attention with Siamese and quadruplet models. 

For a Siamese model, the inputs are pairs $(\mathbf{x}^{1},\mathbf{x}^{2})$. Given their feature vectors $\mathbf{f}^{1}$ and $\mathbf{f}^{2}$, we compute the weight vector $\mathbf{w}$ in the same way as the triplet scenario. If $\mathbf{x}^{1}$ and $\mathbf{x}^{2}$ belong to the same class, $\mathbf{w}=\mathbf{1}-|\mathbf{f}^{1}-\mathbf{f}^{2}|$. If they belong to different classes, $\mathbf{w}=|\mathbf{f}^{1}-\mathbf{f}^{2}|$. With $\mathbf{w}$, we compute the sample scores $s^{1}=\mathbf{w}^{T}\mathbf{f}^{1}$ and $s^{2}=\mathbf{w}^{T}\mathbf{f}^{2}$, and use Equation~\ref{eq:mainSimilarityAttentionMap} to compute attention maps $\mathbf{M}^{1}$ and $\mathbf{M}^{2}$ for $\mathbf{x}^{1}$ and $\mathbf{x}^{2}$ respectively. For a quadruplet model, the inputs are quadruplets $(\mathbf{x}^{a},\mathbf{x}^{p},\mathbf{x}^{n1},\mathbf{x}^{n2})$, where $\mathbf{x}^{p}$ is the positive sample and $\mathbf{x}^{n1}$ and $\mathbf{x}^{n2}$ are negative samples with respect to $\mathbf{x}^{a}$. Here, we compute the three difference feature vectors $\mathbf{f}^{1}=|\mathbf{f}^{a}-\mathbf{f}^{p}|$, $\mathbf{f}^{2}=|\mathbf{f}^{a}-\mathbf{f}^{n1}|$, and $\mathbf{f}^{3}=|\mathbf{f}^{a}-\mathbf{f}^{n2}|$. Following the intuition described in the triplet case, we get the difference weight vectors as $\mathbf{w}^{1}=1-\mathbf{f}^{1}$ for the positive pair and $\mathbf{w}^{2}=\mathbf{f}^{2}$ and $\mathbf{w}^{3}=\mathbf{f}^{3}$ for the two negative pairs. The overall weight vector $\mathbf{w}$ is then computed as the element-wise product of the three individual weight vectors: $\mathbf{w}=\mathbf{w}^{1} \odot \mathbf{w}^{2} \odot \mathbf{w}^{3}$. Given $\mathbf{w}$, we compute the sample scores $s^{a}=\mathbf{w}^{T}\mathbf{f}^{a}$, $s^{p}=\mathbf{w}^{T}\mathbf{f}^{p}$, $s^{n1}=\mathbf{w}^{T}\mathbf{f}^{n1}$, and $s^{n2}=\mathbf{w}^{T}\mathbf{f}^{n2}$, and use Equation~\ref{eq:mainSimilarityAttentionMap} to obtain the four attention maps $\mathbf{M}^{a}$, $\mathbf{M}^{p}$, $\mathbf{M}^{n1}$, and $\mathbf{M}^{n2}$.

\subsection{Learning with similarity mining}
\label{sec:SA_mining}
With our proposed mechanism to compute similarity attention, one can generate attention maps, as illustrated in Figure~\ref{fig:multipleApplications}, to explain why the similarity model predicted that the data sample satisfies the similarity criterion. However, we note all operations leading up to Equation \ref{eq:mainSimilarityAttentionMap}, where we compute the similarity attention, are differentiable and we can use the generated attention maps to further bootstrap the training process. As we show later, this helps improve downstream model performance and generalizability. To this end, we describe a new learning objective, \textsl{similarity mining}, that enables such similarity-attention-driven training of similarity models. 

The goal of similarity mining is to facilitate the complete discovery of local image regions that the model deems necessary to satisfy the similarity criterion. To this end, given the three attention maps $\mathbf{M}^{i}, i\in \{a,p,n\}$ (triplet case), we upsample them to be the same size as the input image and perform soft-masking, producing masked images that exclude pixels corresponding to high-response regions in the attention maps. This is realized as: $\hat{\mathbf{x}}=\mathbf{x} * (\mathbf{1}-\Sigma({\mathbf{M}}))$, where $\Sigma(\mathbf{Z})=\text{sigmoid}(\alpha(\mathbf{Z}-\beta))$ (all element-wise operations and $\alpha$ and $\beta$ are pre-set, by cross validation, constants). These masked images are then fed back to the same encoder of the triplet model to obtain the feature vectors $\mathbf{f}^{*a}$, $\mathbf{f}^{*p}$, and $\mathbf{f}^{*n}$. Our proposed similarity mining loss $L_{sm}$, can then be expressed as:
\begin{equation}
    L_{sm}=\Bigl\lvert\|\mathbf{f}^{*a}-\mathbf{f}^{*p}\|-\|\mathbf{f}^{*a}-\mathbf{f}^{*n}\|\Bigr\rvert
    \label{eq:similarityMining}
\end{equation}
where $\|\mathbf{t}\|$ represents the Euclidean norm of the vector $\mathbf{t}$. The intuition here is that by minimizing $L_{sm}$, the model has difficulties in predicting whether the input triplet would satisfy the triplet condition. This is because as $L_{sm}$ gets smaller, the model will have exhaustively discovered all possible local regions in the triplet, and erasing these regions (via soft-masking above) will leave no relevant features available for the model to predict that the triplet satisfies the criterion.

\subsubsection{Extensions to other architectures}
Like similarity attention, similarity mining is also extensible to other similarity learning architectures. For a Siamese similarity model, we consider only the positive pairs when enforcing the similarity mining objective. Given the two attention maps $\mathbf{M}^{1}$ and $\mathbf{M}^{2}$, we perform the soft-masking operation described above to obtain the masked images, resulting in corresponding features $\mathbf{f}^{*1}$ and $\mathbf{f}^{*2}$. The similarity mining objective then attempts to maximize the distance between $\mathbf{f}^{*1}$ and $\mathbf{f}^{*2}$, \ie, $L_{sm}=-|\mathbf{f}^{*1}-\mathbf{f}^{*2}|$. Like the triplet case, the intuition of $L_{sm}$ here is that it seeks to get the model to a state where after erasing, the model can no longer predict that the data pair belongs to the same class. This is because as $L_{sm}$ gets smaller, the model will have exhaustively discovered all corresponding regions that are responsible for the data pair to be predicted as similar, \ie, low feature space distance), and erasing these regions (via soft-masking) will result in a larger feature space distance between the positive samples. 

For a quadruplet similarity model, using the four attention maps, we compute the feature vectors $\mathbf{f}^{*a}$, $\mathbf{f}^{*p}$, $\mathbf{f}^{*n1}$, and $\mathbf{f}^{*n2}$ using the same masking strategy above. We then consider the two triplets $T_{1}=(\mathbf{f}^{*a},\mathbf{f}^{*p},\mathbf{f}^{*n1})$ and $T_{2}=(\mathbf{f}^{*a},\mathbf{f}^{*p},\mathbf{f}^{*n2})$ in constructing the similarity mining objective as $L_{sm}=L_{sm}^{T_{1}}+L_{sm}^{T_{2}}$, where $L_{sm}^{T_{1}}$ and $L_{sm}^{T_{2}}$ correspond to Equation~\ref{eq:similarityMining} evaluated for $T_{1}$ and $T_{2}$ respectively.

\subsection{Overall training objective}
We train similarity models with both the traditional similarity/metric learning objective $L_{ml}$ (\eg, contrastive, triplet, etc.) as well as our proposed similarity mining objective $L_{sm}$. Our overall training objective $L$ is:

\begin{equation}
    L=L_{ml}+\gamma L_{sm}
    \label{eq:similarityOverall}
\end{equation}
where $\gamma$ is a weight factor controlling the relative importance of $L_{ml}$ and $L_{sm}$. Figure~\ref{fig:SA_mining} provides a visual summary of our training pipeline.

\section{Experiments and Results}
We conduct experiments on three different tasks: image retrieval (Sec.~\ref{sec:retr}), person re-identification (Sec.~\ref{sec:reid}), and one-shot semantic segmentation (Sec.~\ref{sec:segm}) to demonstrate the efficacy and generality of our proposed framework. We use a pretrained ResNet50 as our base architecture and implement all our code in Pytorch.

\subsection{Image Retrieval}
\label{sec:retr}
We conduct experiments on the CUB200 (``CUB") \cite{wah2011caltech}, Cars-196 (``Cars")\cite{krause20133d} and Stanford Online Products (``SOP") \cite{LiftStruct_CVPR15} datasets, following the protocol of Wang \etal \cite{msLoss_CVPR19}, and reporting performance using the standard Recall@K (R-K) metric \cite{msLoss_CVPR19}. We first show ablation results to demonstrate performance gains achieved by the proposed similarity attention and similarity mining techniques. Here, we also empirically evaluate our proposed technique with three different similarity learning architectures to demonstrate its generality. In Table~\ref{table:ablation}, we show both baseline (trained only with $L_{ml}$) and our results with the Siamese, triplet, and quadruplet architectures (trained with $L_{ml}+\gamma L_{sm}$). As can be noted from these numbers, our method consistently improves the baseline performance across all three architectures. Since the triplet model gives the best performance among the three architectures considered in Table~\ref{table:ablation}, for all subsequent experiments, we only report results with the triplet variant. We next compare the performance of our proposed method with competing, state-of-the-art metric learning methods in Table~\ref{table:SOTAML}. We note our proposed method is quite competitive, with R-1 performance improvement of $2.6\%$ on CUB, matching (with DeML) R-1 and slightly better R-2 performance on Cars, and very close R-1 and slightly better R-1k performance (w.r.t. MS \cite{msLoss_CVPR19}) on SOP.

\begin{table}[h!]
\centering

\caption{Ablation study on CUB dataset. All numbers in $\%$.}

\scalebox{0.9}{
\begin{tabular}{c|c|c|c|c}
\hline
Arch. &Type &R-1 & R-2 &R-4 \\
    \hline\hline
    \multirow{2}{*}{Siamese}&Baseline & 65.9&77.5&\textbf{85.8}\\
    &\textbf{SAM} &\textbf{67.7}&\textbf{77.8}&85.5\\
    \hline
    \multirow{2}{*}{Triplet}&Baseline &66.4&78.1&85.6\\
    &\textbf{SAM} &\textbf{68.3}&\textbf{78.9}&\textbf{86.5}\\
    \hline
    \multirow{2}{*}{Quadruplet}&Baseline & 64.7&75.6&\textbf{85.2}\\
    &\textbf{SAM} & \textbf{66.4}&\textbf{77.0}&\textbf{85.2}\\
    \hline
\end{tabular} 
} 

\label{table:ablation}
\end{table}

\begin{table}[h!]
\centering
\caption{Results on CUB, CARS, and SOP. All numbers in $\%$.}
\vspace{-.5em}
\scalebox{0.9}{
		\setlength{\extrarowheight}{.2em}
		\begin{tabular}{p{2.4cm}|cc|cc|cc}
		    \hline
		     & \multicolumn{2}{c|}{CUB} & \multicolumn{2}{c|}{Cars}  & \multicolumn{2}{c}{SOP}\\
			\cline{2-7}
			& R-1 & R-2 & R-1 & R-2 & R-1 & R-1k\\
			\hline\hline
			Lifted \cite{LiftStruct_CVPR15} & 47.2 & 58.9 & 49.0 & 60.3 & 62.1 & 97.4\\
			N-pair \cite{Npairmc_NIPS16} & 51.0 & 63.3 & 71.1 & 79.7 & 67.7 & 97.8\\
			P-NCA \cite{ProxyNCA_ICCV17} & 49.2 & 61.9 & 73.2 & 82.4 & 73.7 & -\\
			HDC \cite{HDC_ICCV2017} & 53.6 & 65.7 & 73.7 & 83.2 & 69.5 & 97.7\\
			BIER \cite{BIER_ICCV17} & 55.3 & 67.2 & 78.0 & 85.8 & 72.7 & 98.0\\
			ABE \cite{AttentionBasedEF_ECCV18} & 58.6 & 69.9 & 82.7 & 88.8 & 74.7 & 98.0 \\
			MS \cite{msLoss_CVPR19} & 65.7 & 77.0 & 84.1 & 90.4 & \textbf{78.2} & 98.7\\ 
			HDML \cite{zheng2019hardness} & 53.7 & 65.7  &79.1 & 89.7 & 68.7 & -\\
			DeML \cite{chen2019hybrid} & 65.4 & 75.3 & 86.3 & 91.2 & 76.1 & 98.1 \\
			GroupLoss \cite{Elezi_2020_ECCV} & 66.9 & 77.1 & \textbf{88.0} & \textbf{92.5} & 76.3 & - \\
			MS+SFT \cite{SFT_eccv2020} & 66.8 & 77.5 & 84.5 & 90.6 & 73.4 & - \\
			$\text{DRO-KL}_{M}$ \cite{qi2019simple_eccv2020} & 67.7 & 78.0 & 86.4 & 91.9 & - & - \\
			\hline
 			\textbf{SAM}  & \textbf{68.3}&\textbf{78.9} & 86.3 & 91.4 & 77.9 & \textbf{98.8}\\
			\hline
	\end{tabular}} 
\vspace{-.5em}
\label{table:SOTAML}
\end{table}

\begin{figure*}[!h]
	\centering
	\subfloat[Triplet attention maps on CUB dataset for model trained on CUB (left) and CARS (right).]{\includegraphics[draft=false,width=1\linewidth]{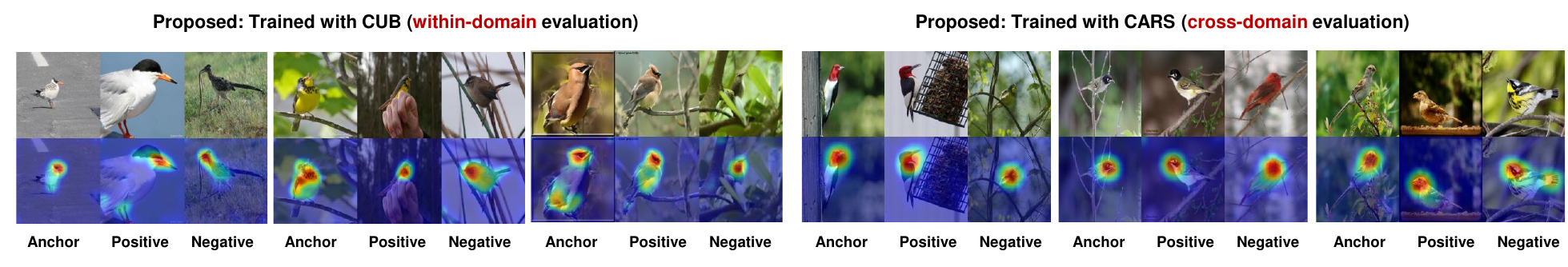}}\\%
	\vspace{-1em}
	\subfloat[Triplet attention maps on CARS dataset for model trained on CARS (left) and CUB (right).]{\includegraphics[draft=false,width=1\linewidth]{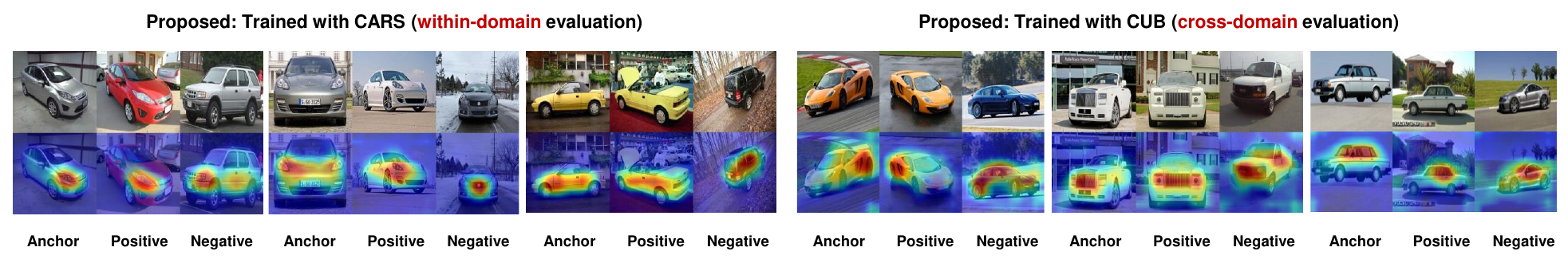}}
	\vspace{-.5em}
	\caption{Triplet attention maps on (a) CUB and (b) CARS with our proposed method for models trained with CUB and CARS.} 
	\vspace{-1em}
	\label{fig:cubcarresults}
\end{figure*}

In addition to obtaining superior quantitative performance, another key difference between our method and competing algorithms is explainability. With our proposed similarity attention mechanism, we can now visualize, by means of similarity attention maps, the model's decision reasoning. In Figures~\ref{fig:cubcarresults}(a) and (b), we show examples of attention maps generated with our method on CUB and Cars testing data with both within- and cross-domain training data respectively. As shown in these figures, our proposed method is generally able to highlight intuitively satisfying correspondence regions across the images in each triplet. For example, in Figure~\ref{fig:cubcarresults}(a) (left 1), the beak color is what makes the second bird image similar, and the third bird image dissimilar, to the first (anchor) bird image. In Figures~\ref{fig:cubcarresults}(a) and (b) (right), we show inter-dataset (cross-domain) results to demonstrate model generalizability. We note that, despite not being trained on relevant data, our model trained with similarity attention is able to discover local regions contributing to the final decision. 

\begin{figure}[h!]
\centering
\includegraphics[width=\linewidth]{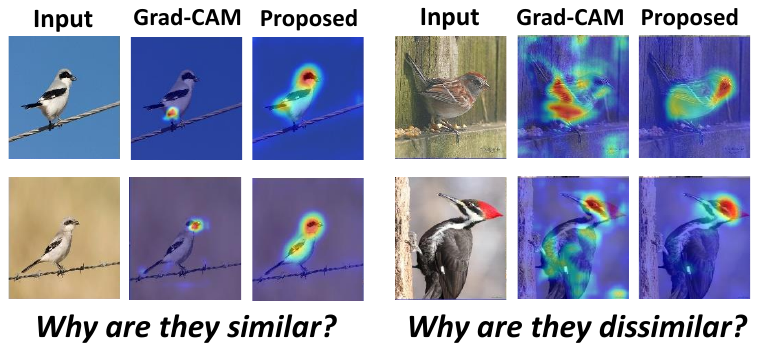}%
\vspace{-.5em}
\caption{\label{fig:gcam} GradCAM (on adapted similarity CNNs with classification head) vs. proposed method. One can note our proposed method is able to highlight corresponding regions more clearly when compared to GradCAM.}
\vspace{-1em}
\end{figure}

While one may add a classification module to any similarity CNN (e.g., Siamese), and then apply GradCAM \cite{GradCAM_ICCV17} to generate class-specific visual explanations, we argue that GradCAM is specifically designed for image classification tasks, which would generate saliency maps for each image (of the input pair) independently and does not ensure faithful explanation of the underlying similarity of the \textit{pair} of images. Concretely, using GradCAM in such an independent fashion may fail to find explicit correspondences between the pair of input images as we shown in Figure \ref{fig:gcam}, since it is designed to highlight regions that contribute to that individual image's classification activations.

\subsection{Person Re-Identification}
\label{sec:reid}
We conduct experiments on person re-id (categorized as a specific image retrieval task) to further prove the efficacy of our proposed framework. We show our proposed  the CUHK03-NP detected (``CUHK") \cite{li2014deepreid,zhong2017re} and DukeMTMC-reid (``Duke") \cite{ristani2016performance,zheng2017unlabeled} datasets, following the protocol in Sun \etal \cite{sunPCB_ECCV18}.

We use the baseline architecture of Sun \etal \cite{sunPCB_ECCV18} and integrate our proposed similarity learning objective of Equation~\ref{eq:similarityOverall}. We set $\gamma=0.2$ and train the model for 40 epochs with the Adam optimizer. We summarize our results in Table~\ref{table:stateofArtReId}, where we note our method results in about $3\%$ rank-1 performance improvement on CUHK and very close performance (88.5\% rank-1) to the best performing method (MGN) on Duke. We note that some of these competing methods have re-id specific design choices (e.g., upright pose assumption for attention consistency in CASN \cite{zheng2019re}, hard attention in HA-CNN \cite{LiHACNN_CVPR18}, attentive feature refinement and alignment in DuATM \cite{si2018dual}). On the other hand, we make no such assumptions, however, is able to achieve competitive performance. 

\begin{table}[!h]
    \vspace{-.5em}
	\centering
	\caption{Re-Id results on CUHK and Duke (numbers in \%).}
	\scalebox{0.95}{
		\begin{tabular}{p{2.8cm}|cc|cc}
		    \hline
		     & \multicolumn{2}{c|}{CUHK} & \multicolumn{2}{c}{Duke} \\
			\cline{2-5}
			& R-1 & mAP & R-1 & mAP \\
			\hline\hline
			SVDNet \cite{SVDnet_ICCV17} & 41.5 &37.3 &76.7 & 56.8\\
			HA-CNN \cite{LiHACNN_CVPR18} & 41.7 & 38.6 & 80.5 & 63.8 \\
            DuATM \cite{si2018dual} & - & - & 81.8&64.6 \\
			PCB+RPP \cite{sunPCB_ECCV18} & 63.7 & 57.5 & 83.3 & 69.2 \\
 			MGN \cite{MGN_MM18}  & 66.8 & 66.0 &\textbf{88.7} & \textbf{78.4}\\
            CASN (PCB) \cite{zheng2019re} & 71.5 & 64.4 & 87.7& 73.7 \\ 
            \hline
            \textbf{Proposed} & \textbf{74.5} & \textbf{67.5} & 88.5 & 75.8 \\
            \hline
	\end{tabular}}
	\vspace{-1em}
\label{table:stateofArtReId}
\end{table} 

\subsection{Weakly supervised one-shot semantic segmentation}
\label{sec:segm}

\begin{figure}[h!]
    \centering
    \includegraphics[width=0.95\linewidth]{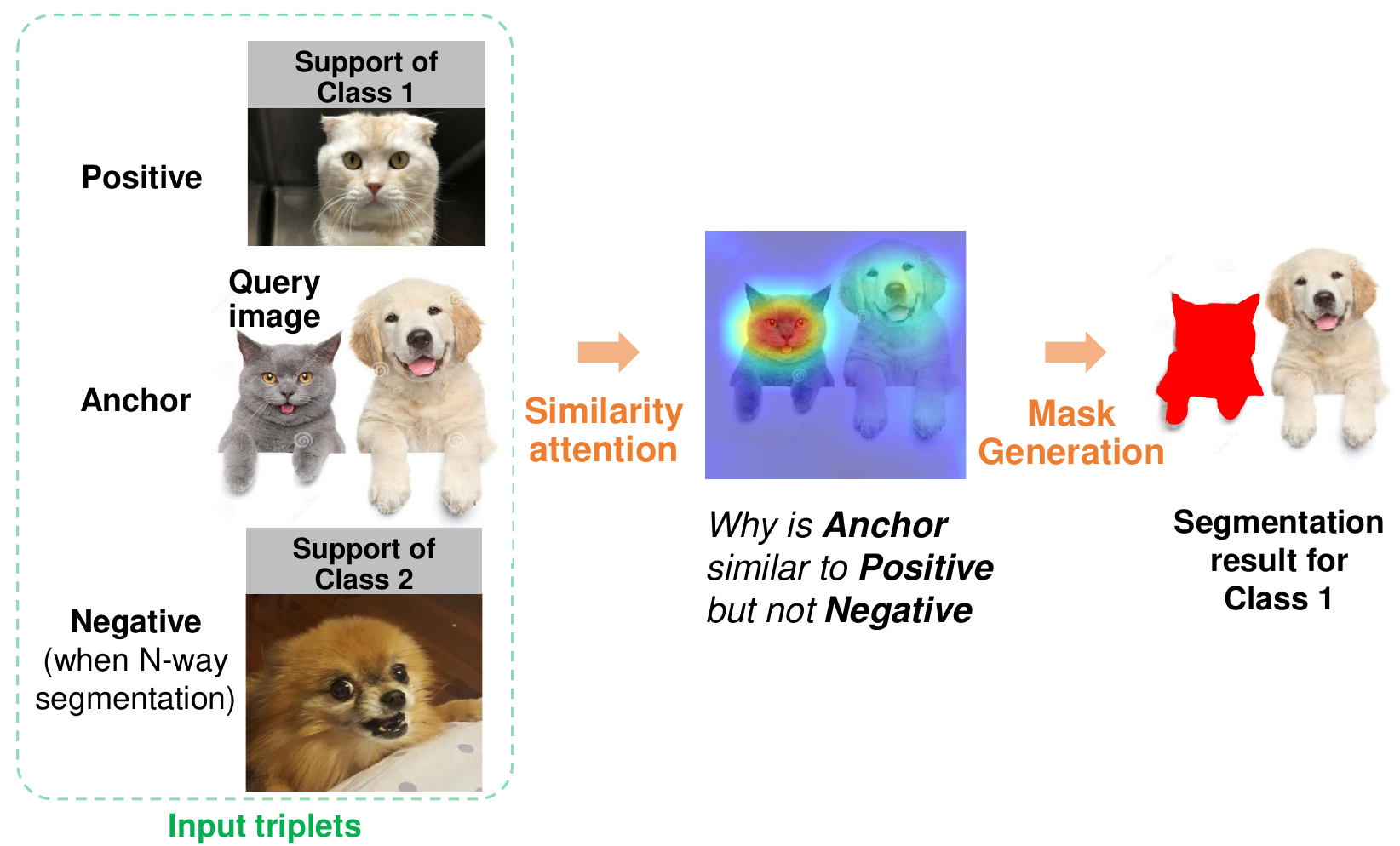}
    \caption{A visual summary of our weakly-supervised one-shot segmentation workflow.}
    \vspace{-0.5em}
    \label{fig:segWorkflow}
\end{figure}

In the one-shot semantic segmentation task, we are given a test image and a pixel-level semantically labeled support image, and we are to semantically segment the test image. Given that we learn similarity predictors, we can use our model to establish correspondences between the test and the support images. With the explainability of our method, the resulting similarity attention maps we generate can be used as cues to perform semantic segmentation. We use the PASCAL$-5^{i}$ dataset (``Pascal") \cite{shaban2017one} for all experiments, following the same protocol as Shaban \etal \cite{shaban2017one}. A visualization of the proposed weakly-supervised one-shot segmentation workflow is shown in Figure \ref{fig:segWorkflow}.

\begin{figure}[h!]
	\centering
	\includegraphics[draft=false,width=\linewidth]{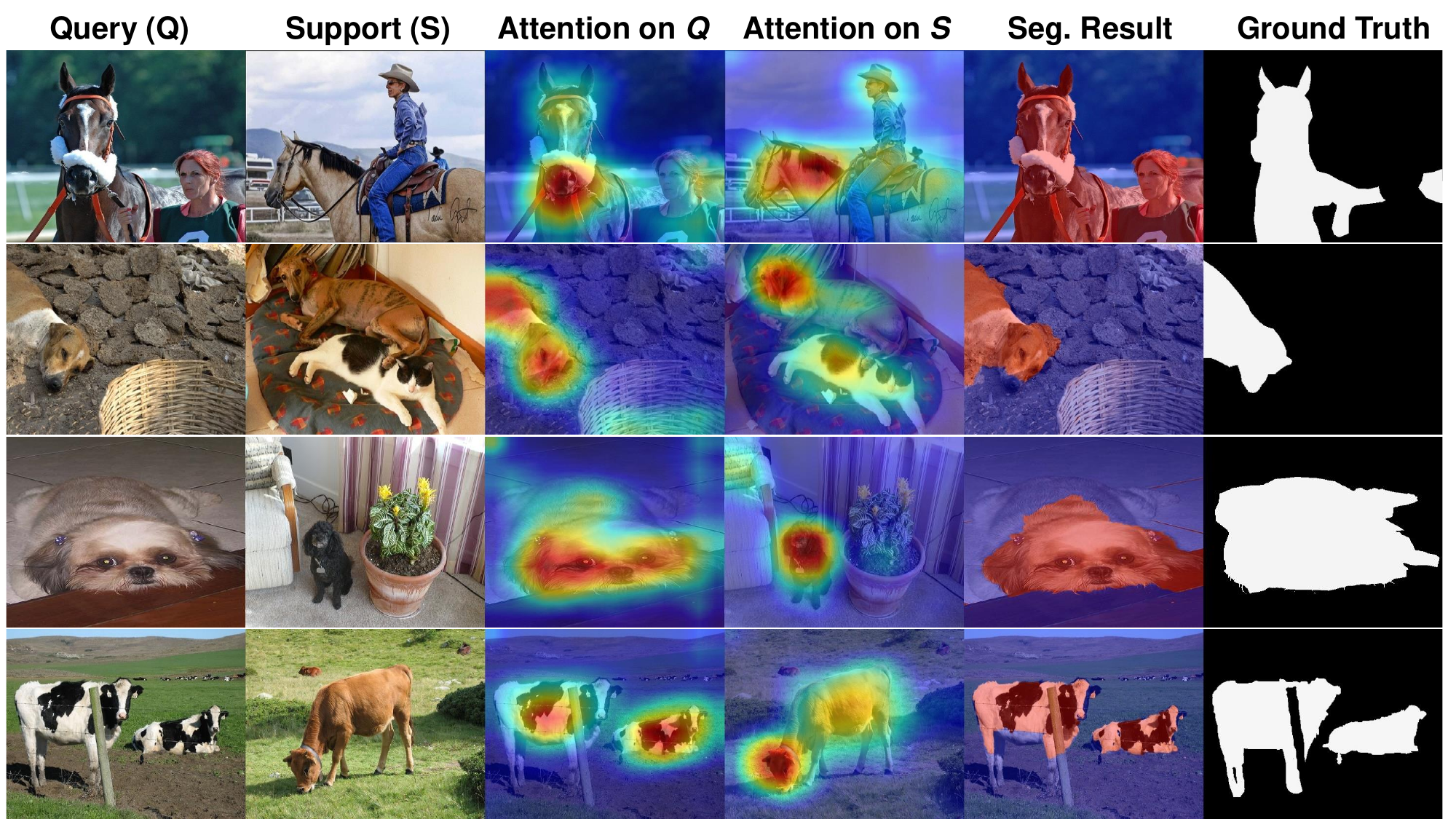}
	\caption{Qualitative one-shot segmentation results from the PASCAL$-5i$ dataset.} 
	\label{fig:oneShotAttResults}
	\vspace{-1.5em}
\end{figure}

\begin{figure*}[h!]
	\centering
	\includegraphics[draft=false,width=0.9\linewidth]{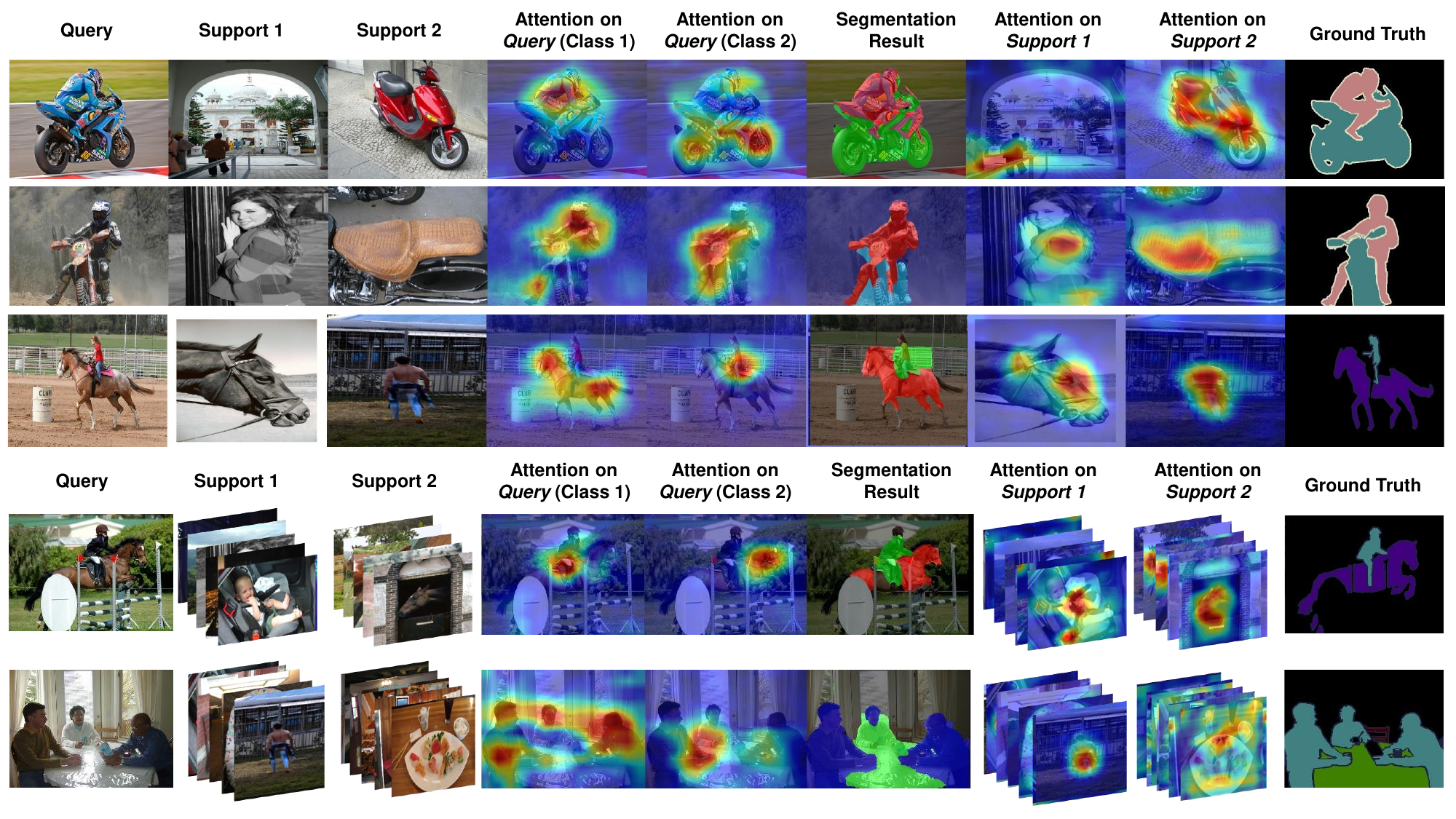}
	\caption{Qualitative one-shot segmentation results from the PASCAL$-5i$ dataset.} 
	\label{fig:2wayAttResults}
	\vspace{-.5em}
\end{figure*}

Given a test image and the corresponding support image, we first use our trained model to generate two similarity attention maps for each image. We then use the attention map for the test image to generate the final segmentation mask using the GrabCut \cite{rother2004grabcut} algorithm. We call this the ``1-way 1-shot" experiment. In the ``2-way 1-shot" experiment, the test image has two objects of different classes and we are given two support images, image 1 and 2 from class 1 and 2 respectively. In this case, to generate results for object class 1, we use the support image 1 as the positive image and support image 2 as negative. Similarly, to generate results for object class 2, we use support image 2 as the positive image and support image 1 as negative. The ``2-way 5-shot" experiment is similar; the only difference is we now have five support images for each of the two classes (instead of one image as above). We first show some qualitative results in Figure~\ref{fig:oneShotAttResults} and \ref{fig:2wayAttResults} (left to right: test image, support image, test attention map, support image attention map, predicted segmentation mask, ground truth mask). In the third row of Figure~\ref{fig:oneShotAttResults}, we see that, in the test attention map, our method is able to capture the ``dog" region in the test image despite the presence of a ``cat" in the support image, helping generate the final segmentation result. In Figure~\ref{fig:2wayAttResults} first row, we see we are able to segment out both the person and the bike following the person and bike categories present in the two support images, helping generate a reasonably accurate final segmentation result. We also show the 1-way and 2-way meanIOU results in Table~\ref{tab:oneShotQuantResults} (following the protocol of \cite{shaban2017one}) and Table~\ref{tab:2wayresults} (following the protocol of \cite{Dong2018FewShotSS}). Here, we highlight several aspects. First, all these competing methods are specifically trained towards the one-shot segmentation task, whereas our model is trained for metric learning. Second, they use the support image label mask both during training and inference, whereas our method does not use this label data. Finally, they are trained on Pascal, \ie, relevant data, whereas our model was trained on CUB and Cars, data that is irrelevant in this context. Despite these seemingly disadvantageous factors, our method performs better than others in some cases and for the overall mean in the 1-way experiment and substantially outperforms competing methods in the 5-way experiments. Finally, we also substantially outperform the recently published PAN-init \cite{PAN_ICCV19} which also does not train on the Pascal data (so this is closer to our experimental setup), while however using the support mask information during inference. These results demonstrate the potential of our proposed method in training similarity predictors that can generalize to data unseen during training and also to tasks for which the models were not originally trained. 

\begin{table}[!h]
\centering
\vspace{-0.5em}
\caption{1-way 1-shot binary-IOU results on PASCAL$-5^{i}$. All numbers in \%.}
\vspace{-0.5em}
\scalebox{0.9}{
\begin{tabular}{c|c|c|c|c|c|c}
    \hline
    Methods & Label &$5^{0}$ &$5^{1}$ & $5^{2}$ & $5^{3}$ & Mean \\
    \cline{1-7}
    \hline\hline
    OSVOS \cite{caelles2017one}& Yes & 24.9 & 38.8 & 36.5 & 30.1 & 32.6 \\
    OSLSM \cite{shaban2017one} & Yes & 33.6 & \textbf{55.3} & 40.9 & 33.5 & 40.8 \\
    co-FCN \cite{rakelly2018conditional} & Yes & 36.7 & 50.6 & \textbf{44.9} & 32.4 & 41.1 \\
    \hline
    PAN-init \cite{PAN_ICCV19} & Yes & 30.8 & 40.7 & 38.3 & 31.4 & 35.3 \\
     \hline
    \textbf{SAM} & \textbf{No} & \textbf{37.9} & 50.3  & 44.4 & \textbf{33.8} & \textbf{41.6} \\ 
    \hline 
\end{tabular}}
\vspace{-1em}
\label{tab:oneShotQuantResults}
\end{table}

\begin{table}[!h]
\centering
\caption{2-way 1-shot and 5-shot binary-IOU results on PASCAL$-5^{i}$. All numbers in \%.}
\vspace{-0.5em}
\scalebox{1}{
\begin{tabular}{c|c|c|c}
      \hline
      Methods & Label & 1-shot & 5-shot \\
      \hline\hline
      PL \cite{Dong2018FewShotSS} & Yes & 39.7 & 40.3 \\ 
      PL+SEG \cite{Dong2018FewShotSS} & Yes & 41.9 & 42.6 \\ 
      PL+SEG+PT \cite{Dong2018FewShotSS} & Yes & 42.7 & 43.7 \\ 
      \hline
      \textbf{SAM} & \textbf{No} & \textbf{56.9} & \textbf{60.1} \\ 
      \hline
\end{tabular}}
\vspace{-1.5em}
\label{tab:2wayresults}
\end{table}

\section{Summary and Future Work}
We presented new techniques to explain and visualize, with gradient-based attention, predictions of similarity models. We showed our resulting \textsl{similarity attention} is generic and applicable to many commonly used similarity architectures. We presented a new paradigm for learning similarity functions with our \textsl{similarity mining} learning objective, resulting in improved downstream model performance. We also demonstrated the versatility of our framework in learning models for a variety of unrelated applications, \eg, image retrieval (including re-id) and low-shot semantic segmentation, where one can easily extend this approach to generate explanations for set-to-set matching problems as well.



\section*{Acknowledgments}
This material is based upon work supported by the U.S. Department of Homeland Security, Science and Technology Directorate, Office of University Programs, under Grant Award 18STEXP00001-03-02, Formerly 2013-ST-061-ED0001. The views and conclusions contained in this document are those of the authors and should not be interpreted as necessarily representing the official policies, either expressed or implied, of the U.S. Department of Homeland Security.

\bibliographystyle{named}
\bibliography{ijcai22}

\end{document}